\documentclass[runningheads]{llncs}
\usepackage[T1]{fontenc}
\usepackage{graphicx}
\usepackage{hyperref}
\hypersetup{colorlinks=true, linkcolor=blue, filecolor=blue, urlcolor=blue, citecolor=blue}
\usepackage{color}

\usepackage{booktabs}
\usepackage{siunitx}
\usepackage{amsmath}
\usepackage{amsfonts}
\usepackage{amssymb}
\usepackage[boxed,lines numbered,no end,inoutnumbered]{algorithm2e}
\usepackage{multicol}
\usepackage{multirow}
\usepackage{array} 
\usepackage[caption=false]{subfig}

\usepackage[inkscapelatex=false]{svg}

\begin{document}
\title{
Scalable Knee-Point Guided Activity Group Selection in Multi-Tree Genetic Programming for Dynamic Multi-Mode Project Scheduling \thanks{This paper has been accepted by the Pacific Rim International Conference Series on Artificial Intelligence (PRICAI) 2025.}
}
\titlerunning{
Scalable Activity Group Selection in Multi-Tree GP for DMRCPSP
}
%
\author{Yuan Tian\inst{1} \and Yi Mei\inst{1} \and
Mengjie Zhang\inst{1}}
\authorrunning{Y. Tian, et al.}
%
\institute{
Centre for Data Science and Artificial Intelligence \& \\School of Engineering and Computer Science, \\Victoria University of Wellington,  Wellington, 6014, New Zealand\\
\email{\{yuan.tian, yi.mei, mengjie.zhang\}@ecs.vuw.ac.nz}}

\maketitle              
\begin{abstract}

The dynamic multi-mode resource-constrained project scheduling problem is a challenging scheduling problem that requires making decisions on both the execution order of activities and their corresponding execution modes. Genetic programming has been widely applied as a hyper-heuristic to evolve priority rules that guide the selection of activity-mode pairs from the current eligible set. Recently, an activity group selection strategy has been proposed to select a subset of activities rather than a single activity at each decision point, allowing for more effective scheduling by considering the interdependence between activities. Although effective in small-scale instances, this strategy suffers from scalability issues when applied to larger problems. In this work, we enhance the scalability of the group selection strategy by introducing a knee-point-based selection mechanism to identify a promising subset of activities before evaluating their combinations. An activity ordering rule is first used to rank all eligible activity-mode pairs, followed by a knee point selection to find the promising pairs. Then, a group selection rule selects the best activity combination. We develop a multi-tree GP framework to evolve both types of rules simultaneously. Experimental results demonstrate that our approach scales well to large instances and outperforms GP with sequential decision-making in most scenarios.

\end{abstract}
\section{Introduction}

The multi-mode resource-constrained project scheduling problem (MRCPSP) \cite{weglarz_project_2011} is a challenging problem in project management, reflecting real-world scenarios where activities can be executed in various modes, each with distinct resource requirements and durations. Its complexity arises from the need to simultaneously satisfy precedence constraints and optimise resource allocation across multiple modes. In real-world applications, project environments are often dynamic and uncertain. Unexpected events, such as changes in resource availability or delays in activity execution, require timely adjustments to the schedule. The dynamic MRCPSP (DMRCPSP) addresses this by allowing real-time adaptations in response to such disturbances. This paper focuses on a variant of DMRCPSP in which activity durations are uncertain.

Dynamic scheduling problems require real-time decision-making based on the ever-changing environment and resource constraints. One common approach to tackling these problems is the use of scheduling heuristics, which calculate priority values for activities in the eligible set and select the activity with the highest priority to execute. While scheduling heuristics can provide reasonable solutions, their design often depends heavily on domain knowledge and requires extensive testing. In contrast, genetic programming (GP) is a powerful hyper-heuristic algorithm that can automatically discover scheduling heuristics through an evolutionary process. Unlike heuristic rules, GP does not rely on prior knowledge but typically evolves effective heuristic rules. This makes GP highly flexible and capable of adapting to different scenarios without the need for manual intervention. GP has been successfully applied to various complex optimization problems \cite{zhang_survey_2023}, demonstrating its potential for solving dynamic scheduling problems.

In existing GP methods \cite{tian_learning_2024,tian_generating_2025} for solving the DMRCPSP, the evolved heuristic rules typically evaluate each activity in the eligible set by computing a priority value and then select the activity-mode pair with the highest priority. The selected activity is executed, the eligible set is updated, and the process is repeated until no activity can be started at the current time. This sequential selection strategy is intuitive but inherently limited as it fails to consider the coordination among multiple activities that could be executed simultaneously as a group. To address this limitation, a group selection strategy \cite{tian_group_2025} has been proposed in previous studies, demonstrating improved performance over sequential selection in small-scale instances. However, this approach suffers from poor scalability, as the number of potential activity groups grows exponentially with the size of the eligible set.

To alleviate the scalability bottleneck inherent in existing group selection strategies, we propose a knee-point-based approach. Specifically, we apply an ordering rule to rank individual activity-mode pairs and then introduce a knee point selection to identify a promising subset. We then generate groups only from this filtered set and apply the group selection rule to select the best combination for execution. This approach substantially mitigates the combinatorial explosion encountered in large eligible sets.

The overall objective of this study is to improve the scalability of GP-based methods that employ group selection for solving the DMRCPSP. The research objectives of this work are as follows:

\begin{itemize}
    \item Design a knee-point-based group selection approach that limits the number of candidate groups while preserving decision quality.
    \item Develop a multi-tree GP method that simultaneously evolves an ordering rule for single activity-mode pairs and a group priority rule for evaluating activity combinations.
    \item Conduct experiments to evaluate the effectiveness of the proposed algorithm and compare it with existing GP methods based on sequential selection.
\end{itemize}
\section{Background}
\subsection{Problem Description}

The DMRCPSP is described as follows: Consider a project composed of a set of activities \(J\), each subject to precedence constraints such that activity \(i\) cannot start until all predecessors \(j \in P_i\) have completed. Every activity \(i\) may be executed in one of several modes \(m \in M_i\), where mode \(m\) incurs a demand \(k_{i,m,r}\) on each renewable resource \(r \in R\) and has an expected duration \(\hat{d}_{i,m}\). In the dynamic setting, the actual duration \(d_{i,m}\) is revealed only upon activity start and is sampled uniformly from the interval \([\,d_{i,m}^{\min},\,d_{i,m}^{\max}\,]\), with the optimistic and pessimistic bounds derived from \(\hat{d}_{i,m}\). At any time \(t\), the aggregate demand of all concurrently executing activities must not exceed the resource capacities \(K_r\). The objective is to construct a feasible schedule that satisfies both precedence and resource constraints while minimising the project makespan.

\begin{figure}[htbp]
  \begin{minipage}[b]{0.4\linewidth}
    \small
        \centering
        \renewcommand{\arraystretch}{0.6}
        \begin{tabular}{cccccc}
            \toprule
            Activity $i$ &  Mode  & $\hat{d}_{i,m}$ & $d^{min}_{i,m}$  & $ d^{max}_{i,m}$ & R1 \\
            \midrule
            1 & 1 & 0 & 0 & 0  & 0\\
            2 & 1 & 5 & 3 & 7  & 10\\
              & 2 & 6 & 5 & 8  & 6\\
            3 & 1 & 4 & 3 & 7  & 7\\
              & 2 & 7 & 6 & 11 & 5\\
            4 & 1 & 4 & 3 & 6  & 9\\
              & 2 & 8 & 7 & 10 & 8\\
            5 & 1 & 4 & 2 & 5  & 7 \\
              & 2 & 6 & 4 & 8  & 4 \\
            6 & 1 & 3 & 2 & 5  & 9 \\
              & 2 & 5 & 4 & 7  & 6 \\ 
            7 & 1 & 0 & 0 & 0 &  0 \\
            \midrule
            \multicolumn{4}{l}{Resource availability} & & 12\\
            \bottomrule
        \end{tabular}
    \newline    
    (a) Activity information
  \end{minipage}
  \hspace{0.05\textwidth}
  \begin{minipage}[b]{0.45\linewidth}
    \centering
    \includegraphics[width=\linewidth]{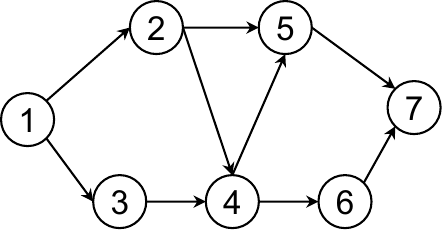}
    (b) Precedence relation
  \end{minipage}
  \caption{Example project.}
  \label{fig:example_project}
\end{figure}

\begin{figure}[htbp]
    \centering
    \subfloat[Example schedule with makespan 20.]{
       \includegraphics[width=0.45\linewidth]{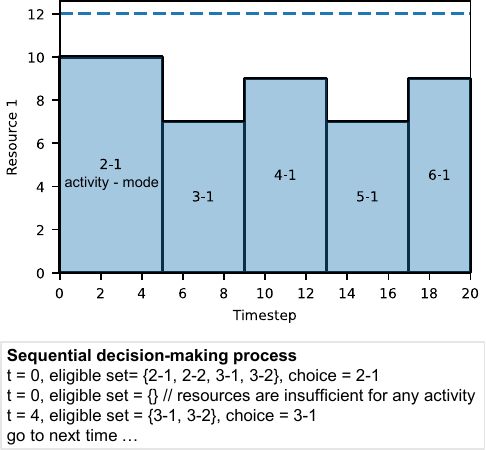}
    }
    \hfill
    \subfloat[Example schedule with makespan 17.]{
        \includegraphics[width=0.45\linewidth]{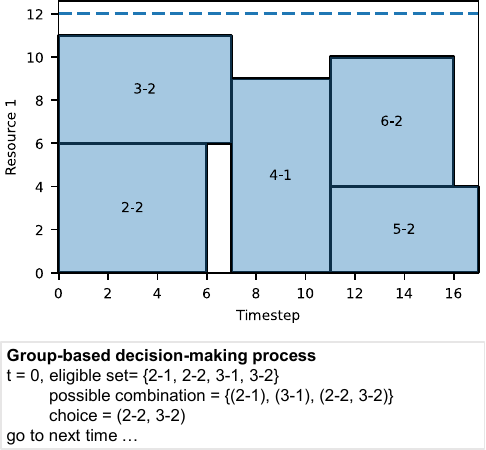}
    }
    \caption{Feasible schedule examples.}
    \label{fig:example_schedule}
\end{figure}

Figure \ref{fig:example_project} illustrates a DMRCPSP instance comprising seven activities, each with two execution modes and a single resource type. Table \ref{fig:example_project}-(a) presents the available modes for each activity, while Figure \ref{fig:example_project}-(b) indicates the precedence relationships. Activities 1 and 7 serve as dummy start and end nodes, respectively. Two feasible schedules are shown in Figure \ref{fig:example_schedule}, where each rectangle represents an activity, with its width corresponding to the duration and height indicating the resource requirement for the given resource type.

\subsection{Genetic Programming Hyper-heuristics for Project Scheduling}

 GP is an effective hyper-heuristics approach to automatically design heuristics for various project scheduling problems, such as single-mode RCPSP \cite{luo_automated_2023}, multi-mode RCPSP \cite{tian_learning_2024,tian_generating_2025} and multi-project RCPSP \cite{chen_guided_2023}. Several techniques, such as surrogate models \cite{luo_automated_2023}, ensemble approaches \cite{dumic_ensembles_2021}, and alternative representations \cite{chen_guided_2023}, have been introduced into the evolutionary process of GP to enhance the effectiveness of the evolved heuristics. 
 
 However, few studies have focused on improving how these heuristics operate within the decision-making process. Most existing approaches adopt, either directly or indirectly, a method known as the parallel schedule generation scheme \cite{kolisch_serial_1996} to rank activities by the priority values evaluated by heuristics and construct schedules step by step. This scheme is characterised by sequentially selecting the activity-mode pairs with the highest priority from the current eligible set. The eligible set is subsequently updated, and the selection process is repeated until no further activities can be scheduled at the current time. This approach is suboptimal as it focuses solely on individual activities, neglecting potential synergies between activity combinations. This limitation becomes more pronounced in the multi-mode RCPSP, where decision-making involves not only selecting activities but also choosing the appropriate execution mode for each activity. A heuritics may favour fast and more resource-intensive modes for individual activities, without considering whether executing more activities simultaneously in slow modes could lead to better overall outcomes.
 
 A group-based selection strategy \cite{tian_group_2025} is proposed to address the aforementioned limitations. The core idea is to enumerate all combinations within the eligible set, evaluate the priority of each group using heuristic rules, and then select the group with the highest priority. The two subfigures in Figure \ref{fig:example_schedule} illustrate the decision-making process and the corresponding makespan of the previously mentioned sequential-style parallel schedule scheme (hereafter referred to as the sequential decision-making strategy) and the proposed group-based selection strategy, respectively. As shown, considering combinations of different activity-mode pairs during decision-making can lead to a better schedule compared to only evaluating the priority of individual activities. However, a major drawback of this approach lies in its combinatorial explosion. Suppose that there are $n$ activities and each activity has $m$ modes, the total number of combinations is $(m+1)^n-1$. If the eligible set contains 10 activities, each with 2 execution modes, the number of possible combinations can reach as high as 50k. Consequently, the approach becomes computationally inefficient when applied to larger-scale projects with more activities in the eligible set. To address this scalability issue, the present work aims to improve the efficiency of the group selection strategy by filtering out unpromising activities in the eligible set before generating candidate groups. This reduction step is expected to limit the number of groups that need to be evaluated, enabling the strategy to scale better in larger problem instances.

\subsection{Knee Point Selection}
Knee point selection \cite{zhang_knee_2015} is commonly used in multi-objective optimisation to identify solutions that offer the most significant trade-offs between conflicting objectives. A knee point on the Pareto front represents a solution where a small improvement in one objective would lead to a substantial deterioration in at least one other objective. This technique has also been applied in GP and shown to be effective, such as in dynamic job shop scheduling \cite{zhang_collaborative_2022,zhang_evolving_2019} and multi-objective GP for feature construction \cite{zhang_improve_2024}, where it helps select promising individuals during evolution. This technique has not been used to select promising activities in the eligible set when solving DMRCPSP. This research introduces this technique into the group selection strategy to further reduce the number of combinations of activities. 

\section{Proposed Method}
\subsection{Overall Framework}

In this paper, we adopt a Koza-style GP framework \cite{koza_genetic_1992}, which is shown in Algorithm \ref{algo:GP_framework}. Each individual is represented using a multi-tree structure \cite{zhang_genetic_2018} that evolves two types of heuristic rules $\mathcal{H}=<\sigma,\gamma>$: an ordering rule $\sigma$ for ranking individual activity-mode rules, and a group priority rule $\gamma$ for evaluating the priority of a group of activities. The terminal and function sets for both trees are detailed in Section \ref{section:terminals}. 

The algorithm begins by randomly generating a set of individuals to form the initial population $pop$. During the evolutionary process (lines 5-10), each individual is evaluated on a set of training instances by constructing schedules and computing its fitness (line 6). Details of the evaluation procedure and how the two rules are applied in the scheduling process can be found in Section \ref{section:evaluation}. Next, the best rule $\mathcal{H}^*$ in the current generation is identified (line 7). Selection is performed using tournament selection (line 8), and offspring are generated by either subtree mutation, subtree crossover or reproduction (line 9). Crossover is applied between the same type of trees from parent individuals, while mutation is performed independently on each tree type. The evolution continues until a termination criterion (e.g., a fixed number of generations) is met, after which the best-performing individual is returned.

\begin{algorithm}[h]
    \small
    \KwIn{A set of training instances $\mathcal{S}=\{\mathcal{I}_1,...,..., \mathcal{I}_T \}$}
    \KwOut{the best evolved rule $\mathcal{H}^*$}
    \textbf{Initialisation:} \\
    \Indp
    $gen \leftarrow 0$, randomly generated population $pop \leftarrow \{\mathcal{H}_1, ..., \mathcal{H}_n\}$ \\
    \Indm
    \While{gen < maxGen}{
        Evaluate fitness of $\mathcal{H} \in pop$ by solving the training set $\mathcal{S}$\\
        $\mathcal{H}^* \leftarrow$ find the best rule in $pop$ \\
        $O$ $\leftarrow$ apply parent selection to $pop$ \\
        $pop$ $\leftarrow$ apply crossover, mutation and reproduction to $O$ \\
     $gen\leftarrow gen+1$   
    }
    \textbf{return} $\mathcal{H}^*$\
    \caption{GP framework.}
    \label{algo:GP_framework}
\end{algorithm}

\subsection{Fitness Evaluation based on Knee-point-based Group Selection of Eligible
Activities}\label{section:evaluation}

 The fitness function, as described in Eq. \ref{eq:fitness}, is defined as the average deviation of the achieved makespan against the lower bound in all training instances. The lower bound of each instance is precalculated by the critical path method \cite{Kelley_critical_1961}, which disregards resource constraints.

\begin{equation}\label{eq:fitness}
   fitness(\mathcal{H}, \mathcal{S})= \frac{1}{|\mathcal{S}|} \sum_{\mathcal{I} \in \mathcal{S}}{\frac{(makespan(solve( \mathcal{H},\mathcal{I}))-lb_{\mathcal{I}})}{lb_{\mathcal{I}}}}
\end{equation}

In Eq. \ref{eq:fitness}, $solve(\mathcal{H}, \mathcal{I})$ refers to the decision-making process responsible for coordinating the project’s execution dynamics. Algorithm \ref{algo:group_parallel_sgs} outlines how these two types of heuristic rule are used to schedule the execution of activities and ultimately complete the project. At time $t=0$, all resources are initialised and available. The algorithm then enters a loop to monitor and update the project’s execution status (lines 4-–9). In each iteration, the current state is updated by checking the progress of the project (e.g., updating completed and ongoing activities) and the availability of resources. The eligible set $E$, which contains all activity-mode pairs that can be started at the current time, is then generated. Using the ordering rule $\sigma$ and group selection rule $\gamma$, the algorithm selects a subset of activity-mode pairs from $E$ to execute, and allocates the required resources. If there are still activities that can be started, the algorithm continues to select and execute them (lines 6--8); otherwise, it advances to the next time unit. Once all activities have been completed, the project is considered finished, and the final schedule is returned.

\begin{algorithm}[h]
    \label{algo:group_parallel_sgs}
    \small
    \KwIn{\text{instance} $\mathcal{I}$; $\mathcal{H}$=<\text{ordering rule} $\sigma$, \text{group priority rule} $\gamma$>}
    \KwOut{A feasible schedule}
    \textbf{Initialisation:}  time: $t\leftarrow0$; all resources are ready \\
    \While{\text{project has not completed}}{
        Update project status and resource availability \\
            \While{$E \leftarrow$ get current eligible set}{
                
                $D \leftarrow kneePointGroupSelection(E, \sigma, \gamma)$ \\
                Start activities $j$ with mode $m$ in $D$
            } 
        
    $t \leftarrow t+1$\\
    }
    \textbf{return }the schedule\\
    \caption{$solve( \mathcal{H},\mathcal{I})$.}
\end{algorithm}

The knee-based group selection (line 7 of Algorithm \ref{algo:group_parallel_sgs}) takes the following five steps to select a subset of activities from the eligible set $E$.

\begin{figure}[t]
    \centering
    \includegraphics[width=.9\linewidth]{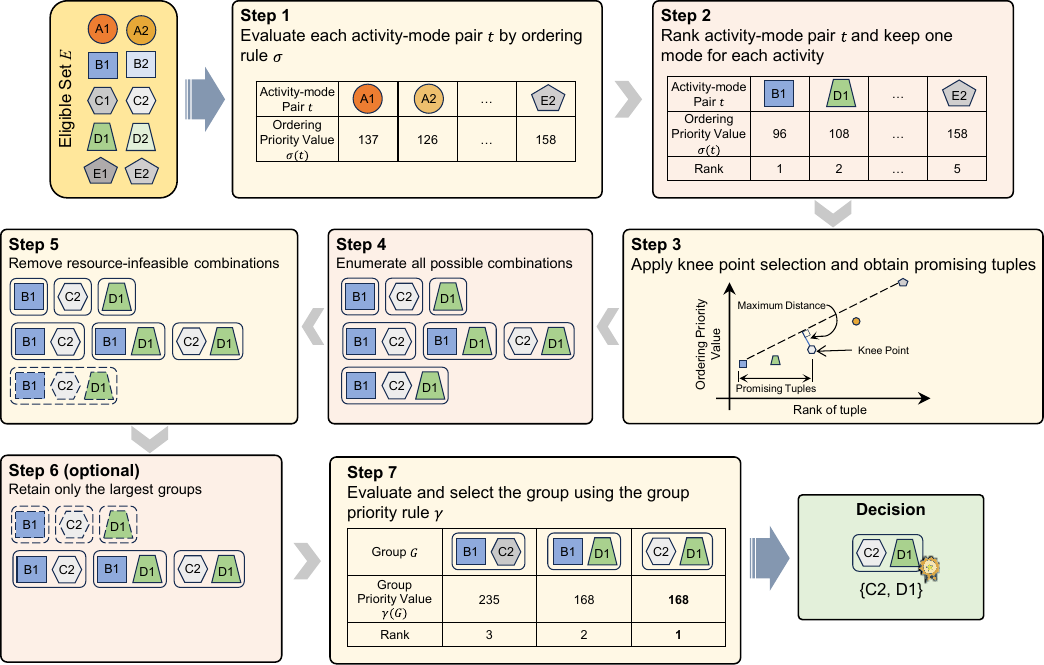}
    \caption{Knee-point based group selection strategy.}
    \label{fig:knee_group_selection}
\end{figure}

\begin{enumerate}
    \item \textbf{Evaluate each activity-mode pair by ordering rule.} Compute the priority value for each activity-mode pair $t$ in the eligible set $E$ using the ordering rule $\sigma$.
    \item \textbf{Rank activity-mode pair and keep one mode for each activity.} Sort all pairs in ascending order according to their priority values and keep one mode with the highest priority for each activity. Note that a smaller priority value indicates a higher priority.
    \item \textbf{Apply knee point selection to obtain promising pairs.} First, form a curve with rank on the x-axis and priority value on the y-axis. Then, construct a straight line connecting the first and the last points of this curve. Next, compute the distance of each point from this line; the point with the maximum distance is identified as the knee point. All pairs with priority values lower than the knee point’s value are selected as the promising pairs. In implementation, if the number of promising activity-mode pairs exceeds 10 after knee point selection, only the top 10 are retained to reduce the enumeration cost.
    \item \textbf{Enumerate all possible combinations.} Enumerate all possible combinations of different sizes from the set of promising pairs.
    \item \textbf{Remove resource-infeasible combinations.} For each combination, calculate its total resource requirement and remove those groups that exceed the current resource constraints.
    \item \textbf{Retain only the largest groups:} Optionally, remove groups that are strict subsets of larger groups. This step reduces the computational load in subsequent evaluations and ensures that the final decision encompasses as many activities as possible, thereby reducing the total number of decisions at the current time. However, this may sometimes reduce the overall decision quality. We treat this step as optional and compare the outcomes in our experiments.
    \item \textbf{Evaluate and select the group using the group priority rule.} Apply the group priority rule $\gamma$ to evaluate each remaining group’s priority, and select the group with the highest group priority value as the final decision.    
\end{enumerate}

Figure \ref{fig:knee_group_selection} illustrates an example of this decision process. Suppose that the eligible set contains five activities (A, B, C, D and E), each available in two modes, yielding ten pairs. In Step 1, all pairs are evaluated and ranked using the ordering rule. In Step 2, these pairs are sorted in ascending order, and the best mode for each activity is kept. In Step 3, the knee point technique selects promising pairs—--activity C in mode 2 is identified as the knee point, and all pairs with priority values lower than that of the knee point are chosen. In Step 4, these promising pairs are enumerated into seven groups; in Step 5, group \{B1, C2, D1\} is discarded because they exceed the current resource constraints. In Step 6, groups that are subsets of larger groups are eliminated. Finally, in Step 7, the remaining three groups are evaluated with the group priority rule, and the final decision is \{C2, D1\}. 

By ranking all activity-mode pairs and selecting only one mode per activity, the method avoids generating multiple combinations for the same activity group with different modes. In addition, the use of the knee-point selection mechanism reduces the number of activities that need to be enumerated. As a result, for an eligible set containing $n$ activities, each with $m$ executable modes, the total number of enumerated combinations is reduced to at most \(2^n - 1\) (with \(n^* < n\)), which partially alleviates the combinatorial explosion issue observed in the previously proposed algorithm \cite{tian_group_2025}.

\subsection{Teriminals and Functions} \label{section:terminals}

Terminals and functions are the fundamental components of GP individuals.
The functions are mathematical operators including $+, -, *, /$(protected), $min$, $max$, $abs$, and $neg$, and take terminals as input arguments. The terminal set, which includes time-related, precedence-related, and resource-related features, is summarised in Table \ref{tab:terminal_set}. These terminals and functions are jointly used to construct both the ordering rule and the group priority rule. 

\begin{table}[t]
\caption{Terminals and their adaptation type for group evaluation.}
\label{tab:terminal_set}
\centering
\scriptsize
\begin{tabular}{l|l|l}
\hline
\multicolumn{1}{c|}{\textbf{Time - Averaging}} & 
\multicolumn{1}{c|}{\textbf{Precedence - Union}} & 
\multicolumn{1}{c}{\textbf{Resource - Aggregation}} \\
\hline
Earliest Start Time & Greatest Rank Positional Weight & Average Resource Requirement \\
Earliest Finish Time & Greatest Rank Positional Weight All & Maximum Resource Requirement \\
Latest Start Time  & Total Predecessor Count & Minimum Resource Requirement  \\
Latest Finish Time  & Direct Predecessor Count & Average Resource Availability  \\
Expected Duration  & Total Successor Count & Maximum Resource Availability \\
Optimistic Duration & Direct Successor Count & Minimum Resource Availability \\
Pessimistic Duration &  & Average Resource Left Afterwards \\
& & Maximum Resource Left Afterwards \\
& & Minimum Resource Left Afterwards \\
& & Resource Required\\
& & Greatest Resource Demand\\
\hline
\end{tabular}
\end{table}
Note that when evaluating a group of activities, the terminals are adapted using three methods \cite{tian_group_2025}: averaging, union, and aggregation. For example, when calculating the earliest start time (EST) of a group such as \{B2, C1\}, the ESTs of both activities are averaged. When computing the direct successor count, the union of each activity's successor set is taken to avoid double-counting. For the average resource requirement, the resource demands of each activity are summed by resource type, and then the average is computed. Additionally, all the time-related terminals are time-invariant \cite{mei_evolving_2017}.

\section{Experimental Study}
\subsection{Experiment Design}
To evaluate the effectiveness of the proposed algorithm, we adopt a simulation model to test the performance of the evolved rules and the decision-making procedure. The simulation model can load various project instances, which we generated based on the parameters listed in Table \ref{tab:project_parameters}. To assess the algorithm under different levels of precedence complexity, number of resource types and resource demand variability, we varied the order strength (OS) and number of resource types $|R|$, resulting in six scenarios. Each scenario is denoted in the form of <OS>/R<$|R|$>. For each scenario, five project instances were generated to maintain evaluation accuracy. The actual activity durations are resampled in each generation during the evaluation phase to prevent overfitting.
\begin{table}[t]
    \caption{Parameter settings of project instance.}
    \centering
    \renewcommand{\arraystretch}{0.9}
    \small
    \begin{tabular}{ll}
        \toprule
         \textbf{Parameter} &  \textbf{Value}\\
         \midrule
         Number of activities & 200 \\
         Number of modes per activity & 3\\
         Activity expected duration & Uniform discrete distribution between 5 and 10\\
         Pessmistic and optimisitc duration & Fluctuate around expected duration by $\pm[1,3]$\\
         Number of resource types & 8, 12\\
         Demands of each resource type & Uniform discrete distribution between 1 and 6\\
         Order strength \cite{demeulemeester_rangen_2003} & 0.75, 0.5, 0.25 \\
         Resource factor \cite{demeulemeester_rangen_2003} & 1 \\
         Resource strength \cite{demeulemeester_rangen_2003} & 0.25\\
         \bottomrule
    \end{tabular}
    \label{tab:project_parameters}
\end{table}

The proposed algorithm is referred to as KGGP. Based on whether Step 6 of the group selection strategy is applied, we further differentiate between KGGP-max (which applies Step 6) and KGGP-all (which skips Step 6). The baseline algorithm is denoted as SGP (sequential-selection GP) \cite{tian_learning_2024}, which employs the parallel schedule generation scheme. The GP parameter settings are the same as the previous research \cite{tian_learning_2024,tian_group_2025}. 


\subsection{Test Performance}
Thirty independent runs are conducted for comparison, and the mean and the standard deviation of the objective value of the three algorithms for six scenarios are shown in Table \ref{tab:objective_values}. The differences between these algorithms are verified by the Wilcoxon rank sum test with a significance level of 0.05. In the following results, "$\uparrow$", "$\downarrow$" and "$\approx$" indicate the corresponding result is significantly better than, worse than or similar to its counterpart. 

\begin{table}[t]
\centering
\caption{ The mean (standard deviation) of the objective values of these algorithms over 30 independent runs for six scenarios.}
\small
\renewcommand{\arraystretch}{0.9}
\begin{tabular}{llll}
\toprule
Scenario & SGP & KGGP-max & KGGP-all \\
\midrule
0.75/R8-1 & 1.641(0.016) & 1.639(0.02)($\approx$) & 1.64(0.018)($\approx$)($\approx$) \\
0.75/R12-1 & 1.751(0.016) & 1.732(0.012)($\uparrow$) & 1.733(0.013)($\uparrow$)($\approx$) \\
0.5/R8-1 & 1.687(0.017) & 1.666(0.014)($\uparrow$) & 1.666(0.019)($\uparrow$)($\approx$) \\
0.5/R12-1 & 1.744(0.02) & 1.706(0.013)($\uparrow$) & 1.708(0.016)($\uparrow$)($\approx$) \\
0.25/R8-1 & 1.669(0.018) & 1.648(0.019)($\uparrow$) & 1.648(0.015)($\uparrow$)($\approx$) \\
0.25/R12-1 & 1.773(0.017) & 1.727(0.025)($\uparrow$) & 1.729(0.02)($\uparrow$)($\approx$) \\
\bottomrule
\end{tabular}
\label{tab:objective_values}
\end{table}

\begin{figure}[t]
    \centering
    \includegraphics[width=0.8\linewidth]{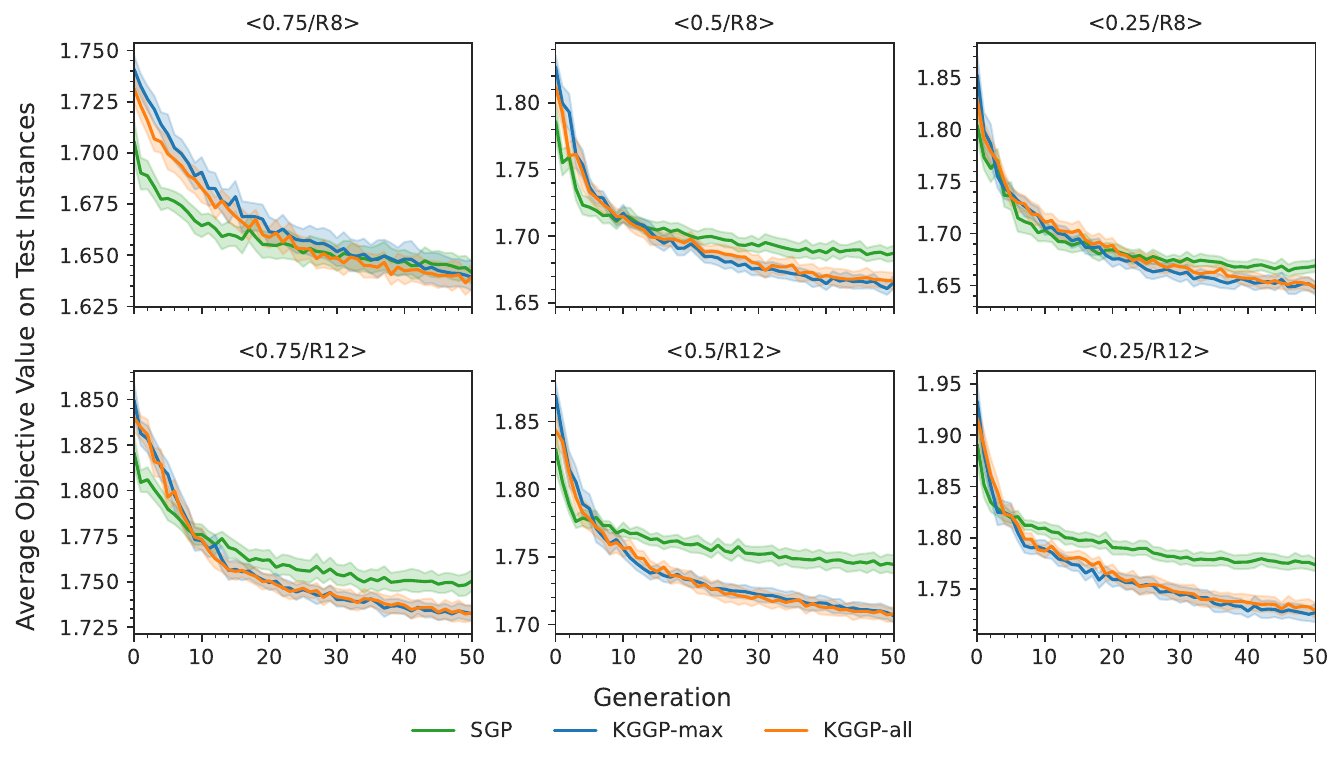}
    \caption{Convergence curves over 30 independent runs in six scenarios.}
    \label{fig:convergence_curve}
\end{figure}

The two KGGP variants outperform the SGP algorithm in the majority of scenarios (5 out of 6). Specifically, the advantage of KGGP over SGP is more pronounced in scenarios with OS = 0.5 and 0.25. This is likely because in such settings, the project contains fewer precedence constraints, allowing more activities to be executed concurrently. As a result, the eligible set at each decision point is larger, making synergistic strategies like group selection more effective. In contrast, when OS is higher, the activity network becomes more constrained, limiting concurrent executions and reducing the size of the eligible set. Under these conditions, the performance difference between group selection and sequential selection becomes less significant.

From the perspective of resource variety, KGGP shows a more substantial improvement over SGP in scenarios with 12 resource types compared with those with 8. This suggests that the group selection strategy offers greater benefits in complex environments with a higher diversity of resource demands.

Figure \ref{fig:convergence_curve} shows the average objective value on test instances of the best-evolved rule over generations. It can be observed that both KGGP variants initially (within the first 10 generations) underperform compared with SGP. This is because KGGP adopts a multi-tree representation, where each individual contains two rules, leading to a larger search space and slower convergence. However, as the number of generations increases, KGGP gradually surpasses SGP in performance. This indicates that the effectiveness of the group selection strategy heavily relies on the quality of the evolved rules. The performance difference between the two KGGP variants is relatively small, suggesting that, at a given time, whether multiple activities are selected in one step to maximise resource usage or scheduled in several smaller steps yields similar results.

\subsection{Rule Size}

\begin{figure}[h]
    \centering
    \includegraphics[width=\linewidth]{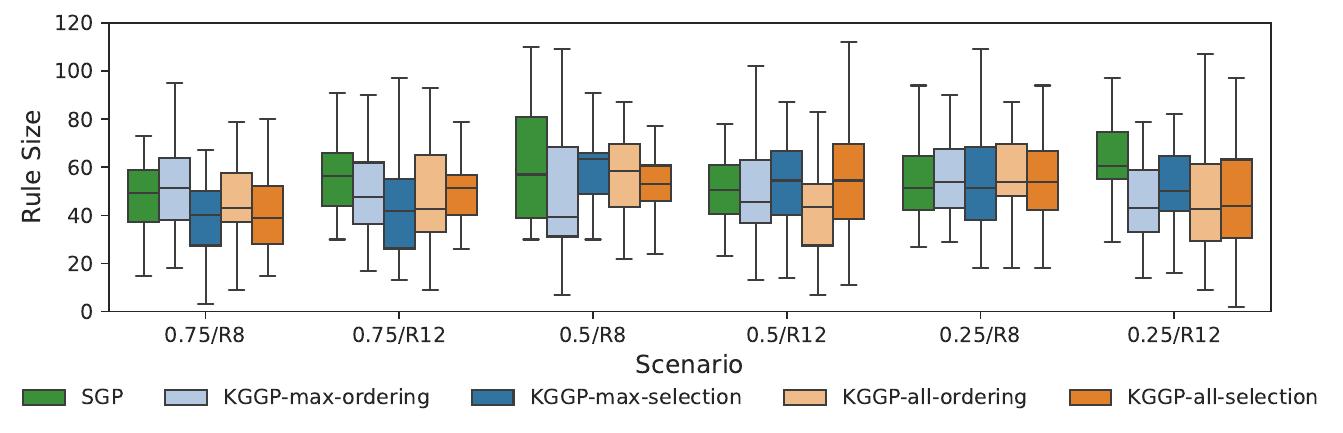}
    \caption{Rule of evolved size in six scenarios over 30 independent runs.}
    \label{fig:rule_size}
\end{figure}

The sizes of the evolved rules generated by these algorithms are shown in Figure \ref{fig:rule_size}. The KGGP variants adopt a multi-tree representation to simultaneously evolve two types of rules: ordering rules and group selection rules. The rule sizes across different algorithms remain generally comparable. This suggests that incorporating group-based decision-making does not necessarily increase rule complexity. 

\subsection{Training Time}

\begin{figure}[ht]
    \centering
    \includegraphics[width=0.8\linewidth]{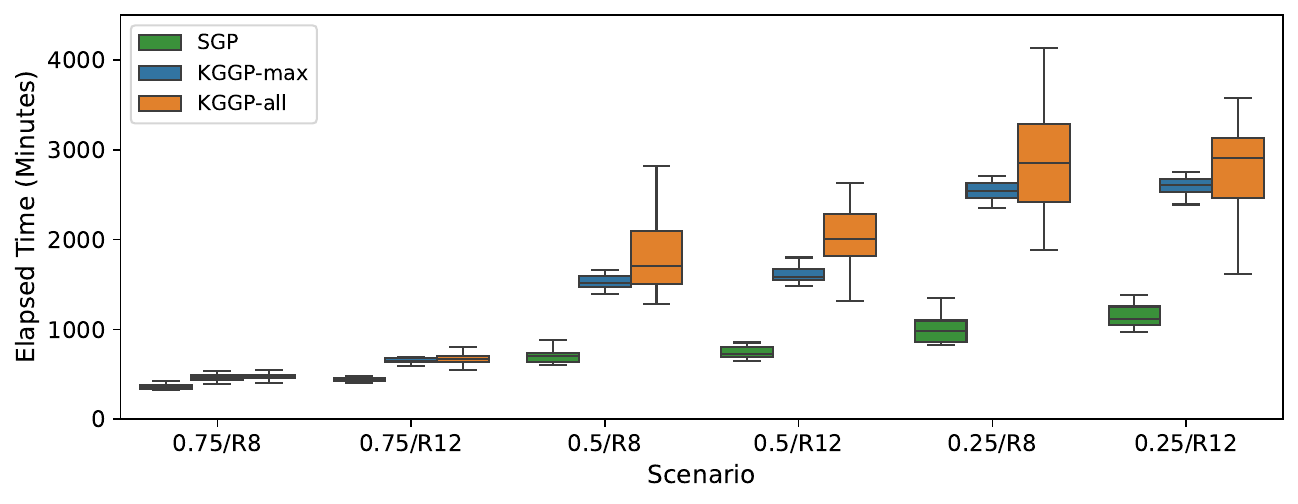}
    \caption{Training time over 30 independent runs in six scenarios.}
    
    \label{fig:training_time}
\end{figure}

The training times of the algorithms across different scenarios are shown in Figure \ref{fig:training_time}. Most of the training time is spent solving project instances using GP individuals, specifically computing the priority values for all pairs. The training time of these algorithms tends to be negatively correlated with the OS value, which reflects the precedence complexity of the project. When the OS value is high, there are more precedence relations among activities, generally resulting in a smaller eligible set and fewer priority value calculations. In contrast, a lower OS value leads to sparser precedence constraints, allowing more activities to be eligible at the same time, which increases the number of pairs to evaluate. 

Within each scenario, KGGP-all takes the longest training time, followed by KGGP-max and SGP. This is because the group selection strategy in KGGP is more complex than the decision strategy used in SGP, requiring more enumeration and evaluation as the size of the eligible set increases. As a result, we observe that in scenarios with OS = 0.75, the training times of the three algorithms are relatively close, while KGGP still outperforms SGP in terms of performance. In scenarios with OS = 0.25, despite yielding better results, KGGP becomes significantly more time-consuming, which suggests a large room for efficiency improvement.  

\subsection{Effectiveness of Knee Point Selection}
\begin{figure}[t]
    \centering
    \includegraphics[width=0.8\linewidth, trim={6pt 6pt 6pt 6pt}, clip]{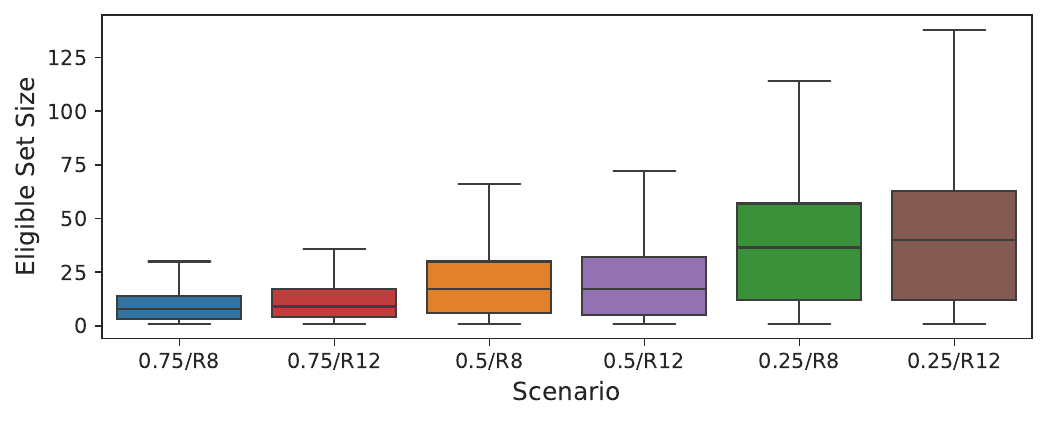}
    \caption{Distribution of eligible set sizes across different scenarios.}
    \label{fig:eligible_size}
\end{figure}

To evaluate the effectiveness of the knee point selection mechanism in reducing the number of candidate combinations, we applied the evolved rules from the KGGP-max algorithm to solve the test instances in each scenario. During the decision-making process, we recorded the size of the eligible set at each decision point, along with the size of the filtered set obtained after applying knee point selection. 
Figure \ref{fig:eligible_size} presents the distribution of eligible set sizes across different scenarios using boxplots. As observed, a decrease in the OS, which corresponds to lower precedence complexity, results in increased scheduling flexibility and thus larger eligible sets. Table \ref{tab:knee_reduction_ratio} summarises the average size of the eligible sets, the average size of the filtered sets after knee point selection, and the corresponding reduction ratios. The results show that knee point selection can effectively eliminate 40\% –- 60\% of the activity-mode pairs, demonstrating its capability to filter out less promising options and provide a more focused candidate set for the subsequent combination enumeration.

\begin{table}[t]
    \centering
    \caption{Average Size of Eligible Sets and Filtered Sets After Knee Point Selection Across Different Scenarios.}
    \small
    \renewcommand{\arraystretch}{0.9}
    \begin{tabular}{lrrr}
    \toprule
    Scenario & 	Eligible Set & Filtered Set & Reduction (\%) \\
    \midrule
    0.75/R8 & 9.70 & 3.74 & 43.85 \\
    0.75/R12 & 11.38 & 4.33 & 45.60 \\
    0.5/R8 & 20.22 & 6.40 & 51.02 \\
    0.5/R12 & 21.39 & 6.91 & 52.03 \\
    0.25/R8 & 38.27 & 11.57 & 56.58 \\
    0.25/R12 & 41.70 & 12.00 & 58.13 \\
    \bottomrule
    \end{tabular}
    \label{tab:knee_reduction_ratio}
\end{table}

\subsection{Rule Analysis of Evolved Scheduling Heuristics}

\begin{figure}[htbp]
    \centering
    \subfloat[Ordering rule]{
       \includegraphics[width=0.25\linewidth]{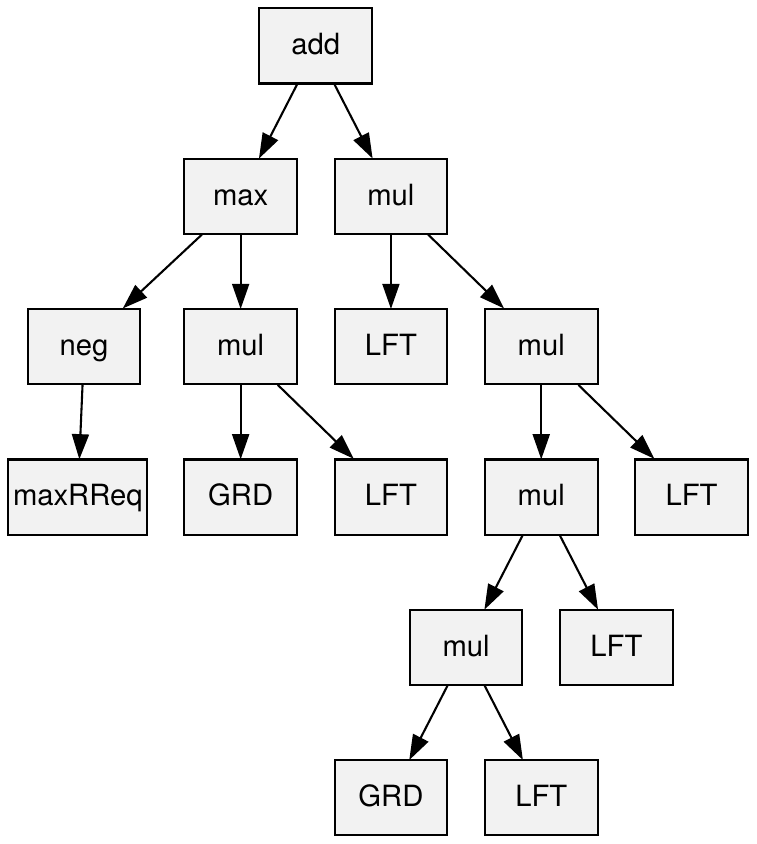}
        \label{fig:ordering}
    }
    \hfill
    \subfloat[Group selection rule]{
        \includegraphics[width=0.7\linewidth]{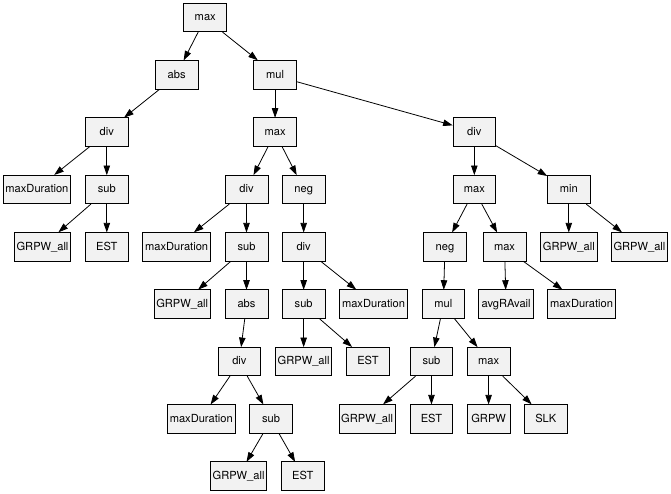}
        \label{fig:group}
    }
    \caption{An evolved GP individual by KGGP-max.}
    \label{fig:example_evolved_rule}
\end{figure}

To examine how the evolved rules operate, we select a rule pair from the 0.5/R8 scenario, visualized in Figure \ref{fig:example_evolved_rule}. The ordering rule emphasizes the term \texttt{GRD * LF}, which appears multiple times, effectively yielding a dominant component proportional to  \texttt{GRD * LF\textsuperscript{4}}. Here, \texttt{GRD} (Greatest Resource Demand) represents the total resource unit-time required by an activity-mode pair, while \texttt{LF} (Latest Finish time) denotes the latest time the activity can finish without delaying the project. This structure favours activities that are both time-critical (small \texttt{LF}) and resource-efficient (small \texttt{GRD}), reinforcing the selection of urgent and efficient activity-mode pairs. The group selection rule primarily involves \texttt{maxDuration} and \texttt{GRPW\_all}, which appear in ratio-based expressions. \texttt{GRPW\_all} (Greatest Rank Positional Weight All) quantifies the total duration of all immediate and indirect successors of the group, reflecting the group’s downstream impact. The rule tends to prefer groups with short execution times and high precedence influence, as indicated by smaller \texttt{maxDuration/GRPW\_all} ratios, corresponding to lower priority values. Overall, these rules reflect a coherent decision-making pattern: (1) The ordering rule selects individual activities that are urgent and cost-effective in resource usage. (2) The group selection rule favours groups that are quick to execute and can unlock a larger portion of the project schedule.

\subsection{Scalability Analysis}
\begin{figure}[h]
    \centering
    \includegraphics[width=1\linewidth]{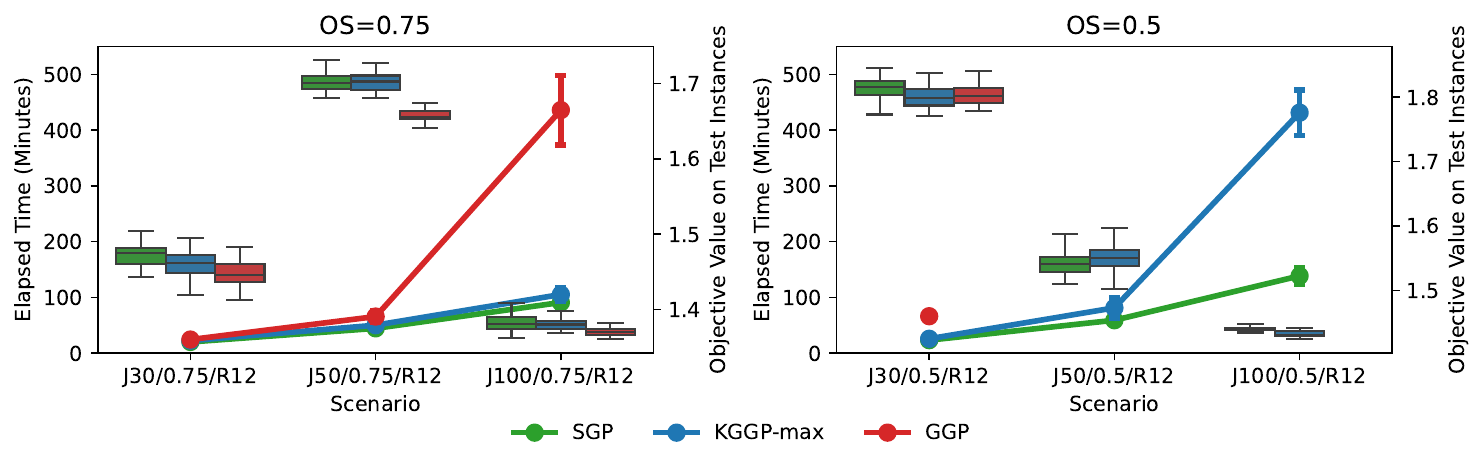}
    \caption{Performance comparison on different project scales and OS levels.}
    \label{fig:scalability_test}
\end{figure}

To evaluate the scalability improvement of our proposed algorithm compared to the existing group-selection-based GP approach (GGP) \cite{tian_group_2025}, we conducted experiments comparing three methods: SGP, GGP, and KGG-max across problem instances of varying sizes and precedence complexity. Specifically, we tested projects with 30, 50, and 100 activities, each having 2 modes, 12 resource types, and OS set to 0.75 and 0.5. Each algorithm was executed 30 times on identical hardware with a wall time limit of 72 hours. Figure \ref{fig:scalability_test} shows the runtime (lineplot, major axis) and the objective values (boxplot, secondary axis) under these scenarios. 

Notably, GGP failed to complete training within the 48-hour limit for both J50 and J100 under OS = 0.5. As expected, increasing the number of project activities under the same OS leads to longer training times for all these algorithms. In the OS = 0.75 setting, the runtime of GGP increases sharply when solving the J100 instances, whereas SGP and KGGP-max exhibit only a moderate increase. This is because, as the number of activities in the project grows, the size of the eligible set expands accordingly. GGP enumerates all activities and their modes within the eligible set, leading to a substantial rise in computational cost. In contrast, KGGP applies a knee point selection strategy to filter out certain activities and retains only one mode per selected activity, thus limiting the growth in computation. In the OS = 0.5 setting, GGP is no longer able to handle the J50 instances, and even KGGP requires a significant amount of time to solve J100. This is because a sparse precedence relationship significantly expands the size of the eligible set as the project scales. For example, in instance J50/0.5/R12, the first decision point involves 12 eligible activities, leading to 531,440 possible combinations. Such large enumeration spaces are computationally expensive. Additionally, we observe that KGGP underperforms GGP in certain scenarios, such as J50/0.75/R12 and J100/0.75/R12. This suggests that although knee-point-based selection reduces the computational burden by limiting candidate combinations, it may also eliminate some potentially high-quality activity-mode pairs. In small-scale instances, full enumeration remains a competitive strategy.
\section{Conclusions and Future Work}
This paper enhances the scalability of the group selection strategy for solving the DMRCPSP by introducing a knee-point group selection mechanism. An activity ordering rule is utilised in the knee-point selection process to identify promising activities, which are then passed to the group selection rule to determine the most suitable activity combination. We developed a multi-tree GP framework to simultaneously evolve both types of rules. Experimental results demonstrate that, under this selection strategy, the evolved rules outperform those generated by sequential activity selection strategies. Furthermore, the knee point selection effectively filters out less-promising activities, making the approach suitable for solving large-scale instances.


Although we have improved the efficiency of group selection by keeping a single mode per activity and considering only a subset of eligible activities, this strategy can still be time-consuming when the eligible set is large. Future work will focus on improving the efficiency of this approach, for example, by adopting heuristic techniques to search for promising combinations instead of enumeration. Additionally, investigating adaptive decision-making strategies that dynamically switch between group selection and sequential selection based on scenario characteristics is another promising direction for future research.


%
%
\bibliographystyle{splncs04}
\bibliography{references}
\end{document}